\documentclass[conference]{IEEEtran}
\IEEEoverridecommandlockouts

\usepackage{amsmath,amssymb}
\usepackage{graphicx}
\usepackage{hyperref}
\usepackage{xcolor}
\usepackage{multirow}
\usepackage{booktabs}
\usepackage{float}
\usepackage{array}
\usepackage{adjustbox}
\usepackage{longtable}
\usepackage{stfloats}
\usepackage{placeins}
\usepackage{titlesec}
\titlespacing{\subsection}{0pt}{12pt plus 2pt minus 2pt}{6pt}

\title{BioMedSearch: A Multi-Source Biomedical Retrieval Framework Based on LLMs}

\author{
    \IEEEauthorblockN{Congying Liu$^{1}$, Xingyuan Wei$^{1,2}$, 
    Peipei Liu$^{1,2}$\IEEEauthorrefmark{2}\thanks{\IEEEauthorrefmark{2} Corresponding Author: peipliu@yeah.net}, 
    Yiqing Shen$^{3}$,
    Yanxu Mao$^{4}$, 
    Tiehan Cui$^{4}$}
    \IEEEauthorblockA{$^{1}$University of Chinese Academy of Sciences, Beijing, China}
    \IEEEauthorblockA{$^{2}$Institute of Information Engineering, Chinese Academy of Sciences, Beijing, China}
    \IEEEauthorblockA{$^{3}$Johns Hopkins University, Baltimore, USA}
    \IEEEauthorblockA{$^{4}$Henan University, Kaifeng, China}
}

\begin{document}
\maketitle
	\begin{abstract}
		Biomedical queries often rely on a deep understanding of specialized knowledge such as gene regulatory mechanisms and pathological processes of diseases. They require detailed analysis of complex physiological processes and effective integration of information from multiple data sources to support accurate retrieval and reasoning. Although large language models (LLMs) perform well in general reasoning tasks, their generated biomedical content often lacks scientific rigor due to the inability to access authoritative biomedical databases and frequently fabricates protein functions, interactions, and structural details that deviate from authentic information. Therefore, we present BioMedSearch, a multi-source biomedical information retrieval framework based on LLMs. The method integrates literature retrieval, protein database and web search access to support accurate and efficient handling of complex biomedical queries. Through sub-queries decomposition, keywords extraction, task graph construction, and multi-source information filtering, BioMedSearch generates high-quality question-answering results. To evaluate the accuracy of question answering, we constructed a multi-level dataset, BioMedMCQs, consisting of 3,000 questions. The dataset covers three levels of reasoning: mechanistic identification, non-adjacent semantic integration, and temporal causal reasoning, and is used to assess the performance of BioMedSearch and other methods on complex QA tasks. Experimental results demonstrate that BioMedSearch consistently improves accuracy over all baseline models across all levels. Specifically, at Level 1, the average accuracy increases from 59.1\% to 91.9\%; at Level 2, it rises from 47.0\% to 81.0\%; and at the most challenging Level 3, the average accuracy improves from 36.3\% to 73.4\%. The code and BioMedMCQs are available at: \url{https://github.com/CyL-ucas/BioMed_Search}
	\end{abstract}
	\section{Introduction}
	
	In recent years, large language models (LLMs), such as DeepSeek~\cite{guo2025deepseek} and ChatGPT~\cite{openai2025gpt4.1}, have achieved remarkable progress in natural language processing tasks. Specifically, they have demonstrated capabilities approaching those of human experts in basic knowledge-based question answering, causal reasoning, and context-dependent instruction comprehension~\cite{yue2025survey}. However, when applied to the biomedical domain, where tasks often involve modeling protein structure–function relationships, analyzing complex disease mechanisms, and predicting biomolecular functions, LLMs are prone to "hallucinations". They may fabricate research conclusions and data sources, or mistakenly associate physiological processes with disease mechanisms~\cite{pal2023medhalt,kim2025medicalhalluc}. Such factually incorrect content not only misleads researchers but also severely impairs their judgment in real-world scientific applications~\cite{xiong2025reliablescientifichypothesisgeneration}. Therefore, integrating authoritative knowledge sources with LLMs to enable dynamic retrieval and rigorous reasoning remains an urgent and open research challenge.
	
	To address this problem, researchers have proposed retrieval-augmented frameworks that integrate external knowledge sources to mitigate hallucinations in LLMs within biomedical scenarios~\cite{cheng2025survey}. These methods enhance semantic reasoning, knowledge attribution, and logical coherence by dynamically retrieving biomedical information from authoritative sources, such as published scientific literature~\cite{guu2020retrieval}. Search agents such as PaSa~\cite{he2024pasa}, MindSearch~\cite{chen2024mindsearch}, and DeepSearcher~\cite{zheng2025deepresearcher} utilize literature databases and web search engines like Google, arXiv, and Brave Search to retrieve relevant documents or web content. They then employ multi-stage filtering based on semantic embedding similarity, summary quality, and contextual relevance to extract high-quality text fragments, which are subsequently integrated into the input of LLMs to enhance the accuracy and traceability of biomedical answer generation. However, these methods still exhibit limited generalization when retrieving protein function and sequence information, often encountering retrieval mismatches. For instance, they may retrieve irrelevant proteins with identical names, leading to erroneous or misleading conclusions generated by the model.
	
	To address the errors encountered by existing methods in specific biomedical tasks, such as mismatches in protein information retrieval,  frameworks like Self‑BioRAG~\cite{jeong2024improving} and MedRAG~\cite{zhao2025medrag} integrate the retrieval augmented generation (RAG) methods with domain-specific knowledge, enhancing the quality of question answering in specialized tasks such as protein-related research and clinical diagnosis through dynamic retrieval, self-reflection mechanisms, knowledge graph integration, and multi-strategy literature filtering. However, such methods typically lack real-time access to web-based information, which may result in missing the latest research developments or failing to capture patients’ authentic feedback and experience sharing regarding treatment processes, recovery progression, or medication responses, leading to incomplete information coverage. Moreover, these methods are primarily evaluated on general biomedical datasets such as MedQA~\cite{jin2021disease}, MedMCQA~\cite{pal2022medmcqa}, and MMLU~\cite{hendrycks2020measuring}, restricting their applicability to specific biomedical scenarios and making it difficult to generalize to more diverse and open-ended contexts.
	
	To this end, we propose BioMedSearch, a novel biomedical retrieval framework that integrates web, literature, and protein databases to achieve more comprehensive and accurate retrieval without any additional training. Our method begins by decomposing the user input into sub-queries and extracting keywords. Subsequently, multi-source retrieval is conducted across literature databases, web search engines, and protein knowledge bases to identify results that effectively address each sub-query, followed by the generation of a comprehensive summary report. To evaluate the retrieval accuracy and reasoning capability of this method, we construct the BioMedMCQs (Biomedical Multi-level Reasoning Multiple-Choice Questions) dataset. It spans three levels of task complexity and enables assessment of both the effectiveness of retrieved information and the model’s ability to perform question answering across different levels of biomedical reasoning.
	Our contributions are as follows:
	\begin{itemize}
		
		\item We propose BioMedSearch, a multimodal search agent for biomedical queries. It supports real-time online retrieval by integrating protein databases, biomedical knowledge bases, and general web search capabilities, ensuring comprehensive and precise access to domain-specific information.
		
		\item We introduce a biomedical search planning component and a retrieval executor. The planner further decomposes sub-queries into fine-grained keywords and organizes their optimal retrieval paths. And the biomedical retrieval executor performs multimodal retrieval across general literature and web sources, protein information databases, and specialized biomedical literature repositories based on the planned paths.
		\item We introduce BioMedMCQs, the first benchmark for biomedical search queries, designed to simulate real-world user intents through randomized generation of biomedical research topics. Tailored for biomedical search-oriented multiple-choice QA, it comprises 3,000 questions spanning three difficulty levels, and is designed to evaluate methods in terms of retrieval effectiveness, biomedical reasoning, and evidence–answer alignment.
	\end{itemize}
	\section{Related Work}
	\subsection{LLM-Based Search Agents for General Information Tasks}
	In recent years, LLM-driven search agents have made significant progress in enhancing retrieval accuracy and content coherence~\cite{zhang2025survey,lewis2021rag}. Representative examples include PaSa~\cite{he2024pasa}, MindSearch~\cite{chen2024mindsearch}, and DeepSearcher~\cite{zheng2025deepresearcher}. PaSa~\cite{he2024pasa} optimizes query rewriting and search navigation via reinforcement learning, MindSearch~\cite{chen2024mindsearch} adopts a multi-agent collaborative framework to simulate human cognitive structures, while DeepSearcher~\cite{zheng2025deepresearcher} introduces an iterative “Read–Search–Reason” loop for dynamic retrieval and structured report generation. These methods improve complex retrieval tasks through task decomposition, embedding-based retrieval, and answer synthesis. However, they rely on general-purpose knowledge bases and web search engines, lacking integration with authoritative biomedical databases such as UniProt and PubMed, which limits their effectiveness in specialized biomedical scenarios. To address this, we design a retrieval framework tailored for biomedical tasks, capable of identifying query intents related to protein structures, gene functions, and disease mechanisms. By integrating general retrieval strategies with domain-specific databases, our approach enhances both generalization and precision in biomedical information retrieval.
	
	\subsection{Domain-Specific Biomedical RAG}
	In biomedical domains, LLMs are prone to hallucinations when generating factual content such as gene-disease associations, protein functions, or treatment outcomes~\cite{pal2023medhalt}. To improve factual accuracy, frameworks like Self-BioRAG~\cite{jeong2024improving} and MedRAG~\cite{zhao2025medrag} integrate domain-specific retrievers, self-reflection mechanisms, and knowledge graphs. Self-BioRAG establishes a retrieval-decision loop with evidence assessment and answer generation, while MedRAG combines sparse and dense retrievers with reasoning chain reranking to enhance coverage and coherence. BiomedRAG~\cite{li2025biomedrag} focuses on general biomedical QA, generating chain-style answers through a lightweight retriever and LLM~\cite{pham2024reliablemedicalquestionanswering}.Despite these improvements, existing biomedical RAG methods typically lack real-time web access and are evaluated on standard datasets like MedQA~\cite{jin2021disease}, MedMCQA~\cite{pal2022medmcqa}, and MMLU~\cite{hendrycks2020measuring}, limiting their adaptability to diverse real-world scenarios and making fine-grained retrieval precision difficult to assess.
	
	In contrast, BioMedSearch integrates structured document retrieval with real-time web and protein database access, enabling dynamic discovery of the latest research and expert opinions. Additionally, we propose a “sub-query decomposition, multi-source filtering, and answerability verification” strategy to ensure the relevance of retrieved content. On our self-constructed BioMedMCQs multi-level evaluation dataset, we empirically demonstrate the advantages of BioMedSearch in medical reasoning and information utilization.
	\section{Method}
	\subsection{Biomedical Search Planner}
	In real-world biomedical query scenarios, users often pose natural language questions such as “Effects of \emph{cyp17a1} knockout(KO) on zebrafish?” While seemingly straightforward, such queries in fact span multiple intersecting concepts~\cite{lin2023decomposing}, including gene function, tissue development, and phenotypic changes. The user may simultaneously seek information on the molecular function of the \emph{cyp17a1} gene, the phenotypic consequences of its KO in zebrafish, its impact on reproductive system development, and even related signaling pathways or 3D protein structures. However, these underlying intents are rarely made explicit in the query, and the required information is scattered across diverse data sources such as literature databases and UniProt. This fragmentation presents significant challenges for automated retrieval framework in accurately identifying user needs and selecting appropriate search tools~\cite{qu2019user,zhang2017bringing}.
	
	To address these complex and task-ambiguous biomedical queries, BioMedSearch introduces a biomedical search planner that sequentially decomposes user query into sub-queries and their corresponding keywords. Based on this, it constructs a directed acyclic graph (DAG) to derive multi-source retrieval paths and allocate appropriate retrieval tools, enabling an efficient transformation from natural language input to executable retrieval workflows.
	
	Specifically, when a user submits a query, the planner first leverages the biomedical knowledge of LLM along with a task-specific decomposition prompt to partition the query into n bounded sub-queries across multiple dimensions, such as developmental effects, physiological or endocrine regulation, clinical phenotypes, and molecular mechanisms. The formal definition of this process is as follows:
	\begin{equation}
		S = \{q_1, q_2, \dots, q_n\} = M(Q, P_{\text{decompose}}, D)
		\label{eq:decompose}
	\end{equation}
	where $Q$ is the original query, $P_{\text{decompose}}$ is the decomposition prompt, $D = \{d_1, d_2, \dots, d_m\}$ denotes the set of biomedical semantic dimensions, M is the language model used for decomposition, and $S = \{q_1, \dots, q_n\}$ is the resulting set of sub-queries. 
	
	To enhance the semantic focus and retrieval adaptability of each sub-query, the planner decomposes each sub-query $q_i$ into $e$ biomedical keywords along fine-grained semantic dimensions such as gonadal development and hormone synthesis. The complete set of extracted keywords is then used to construct a DAG, which serves to clarify the retrieval paths and the allocation of appropriate retrieval tools for each sub-query. This process is formally defined as follows:
	
	\begin{equation}
		\left\{
		\begin{aligned}
			K_i &= \{k_{i1}, k_{i2}, \dots, k_{ie}\} = M_{\text{extract}}(q_i, P_{\text{keywords}}, D') \\
			G &= (V, E) = M_{\text{DAG}}\left( \bigcup_{i=1}^{n} K_i,\ P_{\text{graph}} \right)
		\end{aligned}
		\right.
		\label{eq:keywords_dag}
	\end{equation}
	Here, $K_i$ denotes the set of keywords extracted from the $i$-th sub-query $q_i$; $M_{\text{extract}}$ is the keywords extraction model; $P_{\text{keywords}}$ is the prompt used for keywords extraction; and $D'$ represents the semantic dimensions to which the keywords belong. $G = (V, E)$ denotes the resulting DAG, where the nodes $V$ are derived from the union of all keyword sets $\bigcup_{i=1}^{n} K_i$. 
	
	In summary, biomedical search planner, as the core components of semantic analysis in BioMedSearch, not only enhance the specificity and precision of retrieval but also provide a clear structural foundation for multi-retrieval scheduling, knowledge integration, and report generation.
	\begin{figure*}[t]
		\centering
		\includegraphics[width=\textwidth]{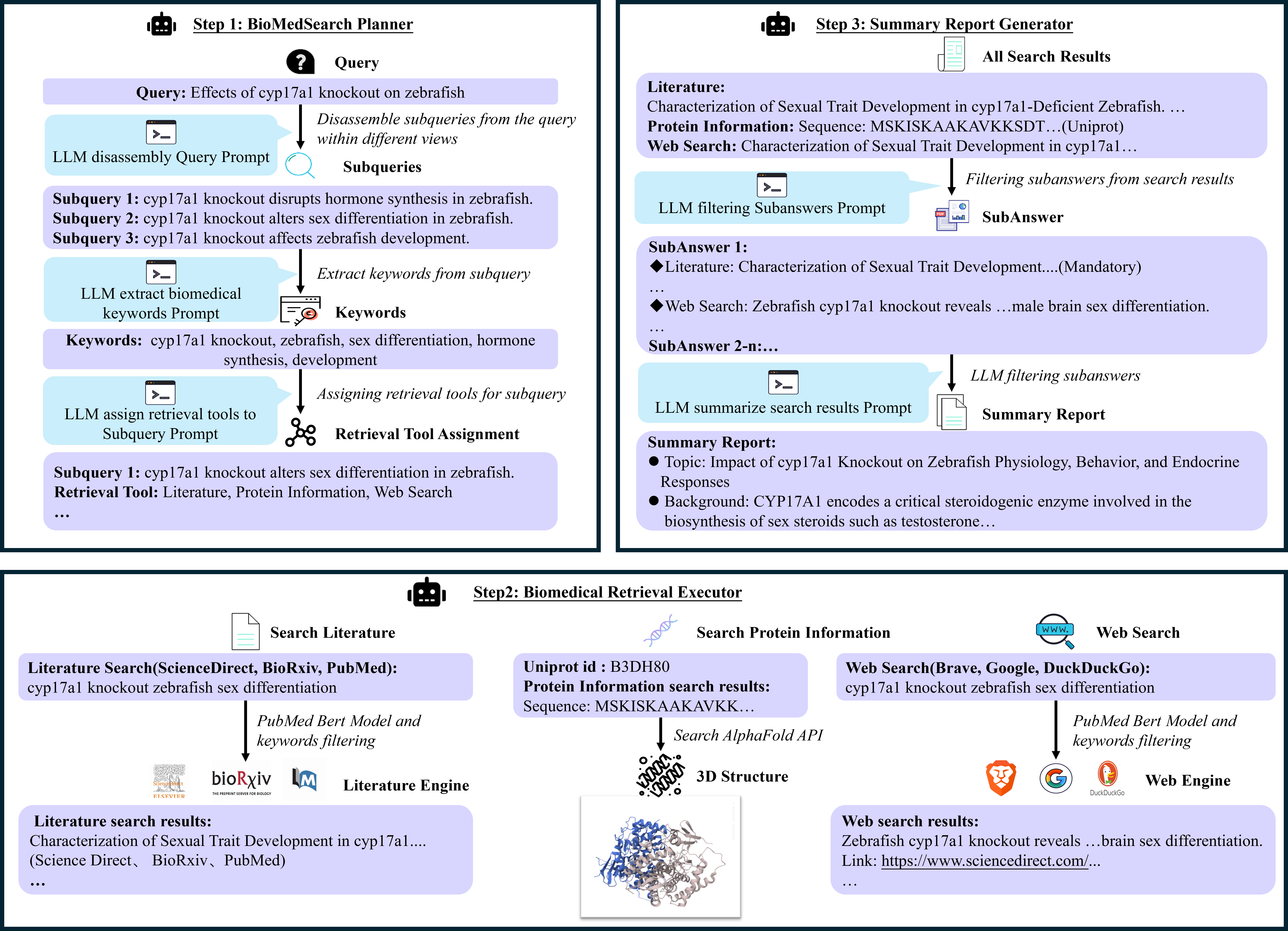}
		\vspace{-5mm}
		\caption{Demonstration of the BioMedSearch workflow using an example query. The purple boxes represent the simulated user input and its corresponding flowchart, including the decomposition of the user query into sub-queries, further keywords extraction, and the planning of appropriate retrieval paths based on these keywords. Subsequently, information retrieval and filtering are performed to obtain relevant results, which are then used to select suitable sub-answers and generate a structured report for user feedback. The blue boxes indicate the core prompting strategies designed for the LLM at each stage of the workflow.}
		\vspace{-5mm}
		\label{fig:get_report_all}
	\end{figure*}
	\subsection{Biomedical Retrieval Executor}
	The Biomedical Retrieval Executor performs multi-source retrieval guided by a DAG-based planning path, which is constructed from extracted keywords and corresponding tool assignments. It consists of three modules: Literature Retrieval, Protein Information Retrieval, and Web Search.
	
	Literature Retrieval: To avoid issues such as weak thematic relevance, content redundancy, or shallow keyword-level matches in literature retrieval~\cite{ebeid2021medgraph}, the executor adopts a dual-strategy approach that combines information coverage with semantic understanding. This strategy aims to accurately identify high-quality publications that are closely aligned with each biomedical sub-query, provide meaningful inferential value, and support the generation of research reports.
	
	When the LLM determines that a sub-query requires literature retrieval, for example, “\emph{cyp17a1} KO alters sex differentiation in zebrafish”, the executor simultaneously queries three major biomedical databases: PubMed, PMC, and ScienceDirect, retrieving up to 100 results from each. If a database fails to return results, the executor reconstructs the query using the three most salient biomedical keywords extracted from the sub-query to ensure broader coverage. Once the papers are collected, the executor applies two consecutive filtering steps to refine semantic alignment and topic relevance. First, it checks keyword coverage by requiring that at least 80\% of the sub-query keywords appear cumulatively in the title and abstract, thereby filtering out documents that are topically irrelevant or only loosely related. Second, it calculates semantic similarity between the sub-query and each paper using vector representations derived from the PubMedBert embedding model, retaining only the top k articles based on cosine similarity.
	
	This integrated filtering process ensures that the final set of retrieved papers is not only topically relevant but also aligned with the user's intent, providing a robust textual foundation for downstream synthesis and research reporting.The overall filtering process can be formally described as follows:
	\begin{equation}
		L^{\text{final}}_i = \text{Top}_k \left( \text{Sim}\left( \text{Filter}_{\geq \theta}(L^{\text{raw}}_i,\ K_i),\ q_i \right) \right)
		\label{eq:literature_filter_refined}
	\end{equation}
	Here, $L^{\text{raw}}_i$ denotes the raw set of literature retrieved for sub-query $q_i$; $K_i = \{k_{i1}, ..., k_{ie}\}$ represents the set of keywords extracted from $q_i$; $\text{Filter}_{\geq \theta}$ is a filtering function that retains only documents with keyword coverage no less than a threshold $\theta$ (empirically set to 80\%); $\text{Sim}(\cdot,\ q_i)$ computes the semantic similarity between the sub-query and each filtered document (typically using cosine similarity); and $\text{Top}_k$ selects the top $k$ most relevant articles (default $k = 10$). The final output $L^{\text{final}}_i$ is the curated literature set most relevant to sub-query $q_i$. Through the integration of multiple literature sources and a semantic-driven filtering framework, the executor effectively identifies and preserves the most representative and high-quality literature, ensuring that complex biomedical queries are addressed with precision, relevance, and interpretability at the retrieval level.
	\enlargethispage{\baselineskip}  
	\begin{figure*}[t]  
		\centering
		\vspace{-3mm}
		\includegraphics[width=\textwidth]{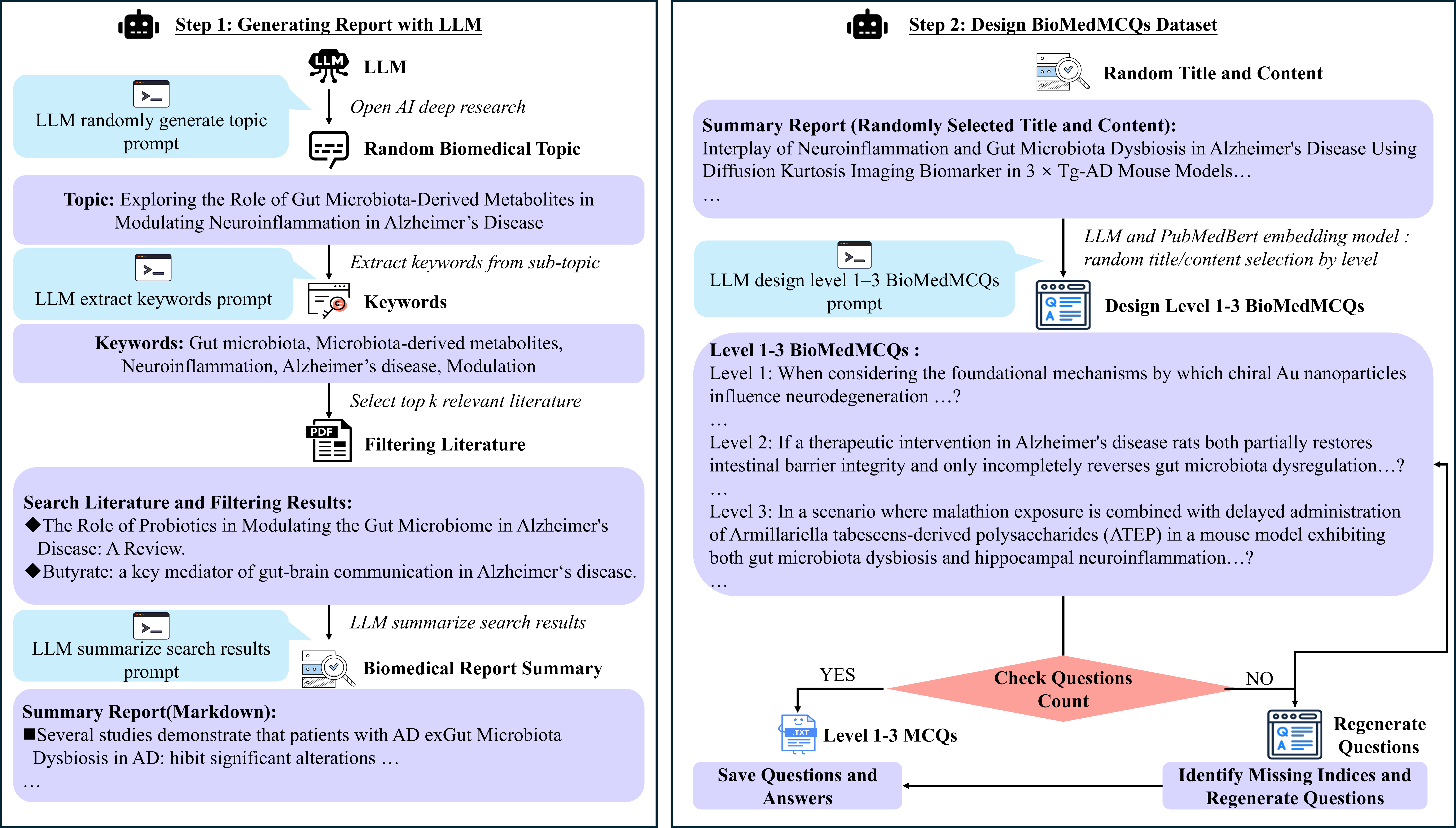}
		\vspace{-3mm}
		\caption{Automated BioMedMCQs generation pipeline based on LLM and literature summarization. This figure illustrates a two-stage pipeline for automatically generating BioMedMCQs using LLM. In the first stage, the LLM generates a random biomedical topic, extracts relevant keywords, retrieves and filters top k literature, and produces a summary report. In the second stage, the report content is used to construct Level 1–3 MCQs, which are iteratively verified and regenerated until the required number of questions is met.}
		\label{fig:design_mcqs_all}
		\vspace{-5mm}
	\end{figure*}
	
	Protein Information Retrieval: The protein information retrieval module in the executor is specifically designed to handle protein-related sub-queries identified by LLM. It leverages contextual understanding to accurately retrieve key biomedical data such as functional annotations and sequence information for the target protein. When an LLM detects the presence of a gene name (with or without an associated organism) in the user input, the framework first performs a UniProt ID lookup. If the sub-query contains descriptive terms such as “effect,” “interaction,” “sequence,” or “protein,” the executor uses the retrieved UniProt ID to extract relevant protein entries from the UniProt database, including molecular function, interactions, and sequence data.
	
	For queries focusing on structural aspects, such as “3D structure” or “spatial conformation”, the executor further invokes the AlphaFold engine, using the UniProt ID to generate and visualize the predicted protein structure in PDB format. This module achieves deep integration between UniProt and AlphaFold, bridging the full spectrum from functional annotation to structural prediction. As a result, it provides robust data support and critical insights for advanced biomedical research, including gene regulatory mechanisms, target identification, and drug design.
	
	Web Search: The is a core component of the executor, designed to effectively handle cases involving non-standard terminology, niche research topics, or cutting-edge developments. By leveraging open search engines, it retrieves real-time and relevant biomedical information and filters for content that is highly aligned with each sub-query. The module integrates three major search engines, namely DuckDuckGo, Google, and Brave, and conducts independent queries for each sub-query. From each engine, it retrieves the top 100 results, including titles, URLs, and summaries. These results are then filtered using a LLM alongside PubMedBert embedding model. The filtering process removes low-quality or irrelevant content, retaining only results with high information density and clear topical focus. The final output is a concise and relevant set of web pages capable of effectively addressing the original sub-query. The filtering Method can be formalized as follows:
	\begin{align}
		R_{\text{Web}}(q) = \text{Top}_k\left( \{ w_i \mid \text{Relevance}(w_i, q) \geq \tau \} \right)
		\label{eq:web_filter}
	\end{align}
	Here, $q$ is the sub-query, and $w_i$ are webpages aggregated from the top 100 results of each search engine. $\text{Relevance}(w_i, q)$ denotes the semantic similarity score between $w_i$ and $q$, computed by the LLM-based filter. $\tau$ is the threshold for filtering, and $\text{Top}_k$ selects the $k$ most relevant webpages to form the final output $R_{\text{Web}}(q)$.
	\subsection{Summary Report Generator}
	After completing all retrieval tasks, the executor proceeds to the result integration and research report generation phase, aiming to transform fragmented information into an accurate research output. Specifically, the framework first employs a LLM to evaluate the retrieval results of each sub-query and extract content that effectively addresses the corresponding question, forming a set of sub-answers. Next, the executor identifies interrelated sub-answers that collectively support the original query intent. These connections serve as the foundation for constructing the final structured research report.
	
	The final generated report encompasses the integrated research background derived from the sub-queries, a synthesis of key findings, source references, and other relevant elements. It ensures the output is well-structured, information-rich, and of practical value for scientific research. Overall, this pipeline leverages a structured and semantically driven design to effectively address the complexity and ambiguity commonly found in biomedical queries. It establishes a clear pathway for downstream retrieval and knowledge integration, while significantly improving the framework’s coherence, precision, and interpretability.
	\section{Experiments}
	\subsection{Dataset and Benchmark Design}
	\begin{figure}[t]
		\centering
		\includegraphics[width=\linewidth]{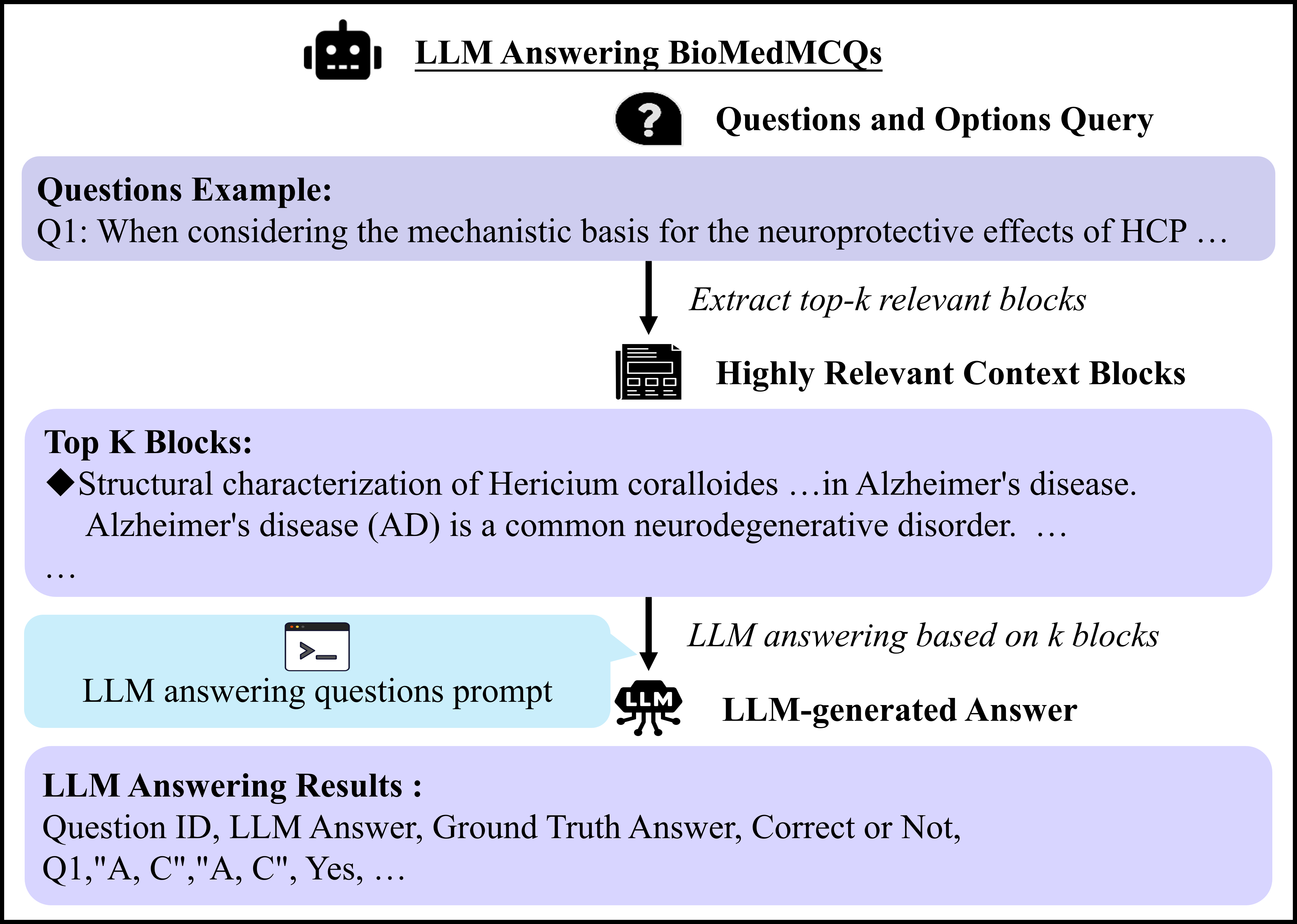}  
		\vspace{-5mm}
		\caption{Illustration of the answer generation on BioMedMCQs using LLMs.}
		\label{fig:Answering_MCQs}
	\end{figure}
	Biomedical multiple-choice questions(MCQs) datasets such as MedQA~\cite{jin2021disease}, MedMCQA ~\cite{pal2022medmcqa}and MMLU~\cite{hendrycks2020measuring} can assess the breadth of LLM’s coverage in clinical or medical domain knowledge and are therefore widely used to evaluate the overall performance of LLMs in MCQs tasks. Although these existing biomedical MCQs datasets have played an important role in testing LLMs’ memory and retrieval capabilities for biomedical knowledge, they are still insufficient in evaluating the accuracy and relevance of retrieved information, as well as the LLM’s ability to understand and reason over such information. Therefore, there is an urgent need to construct a benchmark centered on user input, integrating contextual awareness, reasoning stratification, and cognitive structure modeling, to assess the model’s comprehensive ability to acquire, understand, and apply retrieved information for question answering and the generation of research reports based on retrieved content.
	
	To this end, we propose the BioMedMCQs dataset. This dataset leverages the deep research capability of LLMs to randomly generate a biomedical research topic, simulating user queries to ensure that the evaluation process closely mirrors real-world scientific questioning scenarios. After generating the topic, we further decompose them from biomedical perspective into sub-topics, and then extract keywords from each sub-topic, following the section method. Based on these keywords, we retrieve approximately 600 articles from PMC, PubMed, and ScienceDirect. Subsequently, we filter and select the top 300 articles relevant to each topic through keywords matching and semantic validation using PubMedBert embedding model. Based on this, we construct a multi-dimensional, hierarchical MCQs set covering three representative levels of reasoning complexity with progressively increasing difficulty. Level 1 focuses on fundamental biomedical causal mechanisms, evaluating the LLM’s ability to identify the regulatory role of a single factor; Level 2 emphasizes non-adjacent semantic integration, requiring recognition of implicit logical relationships across sentence boundaries; Level 3 involves hierarchical reasoning under temporal dependency and feedback loops, with contexts encompassing complex clinical or molecular dynamic processes, such as drug metabolism pathways and inflammation regulatory networks. Additionally, each question is manually reviewed by biomedical domain experts to ensure its scientific validity, linguistic clarity, and high relevance to real-world research scenarios.
	\begin{figure*}[b]
		\centering
		\includegraphics[width=\textwidth]{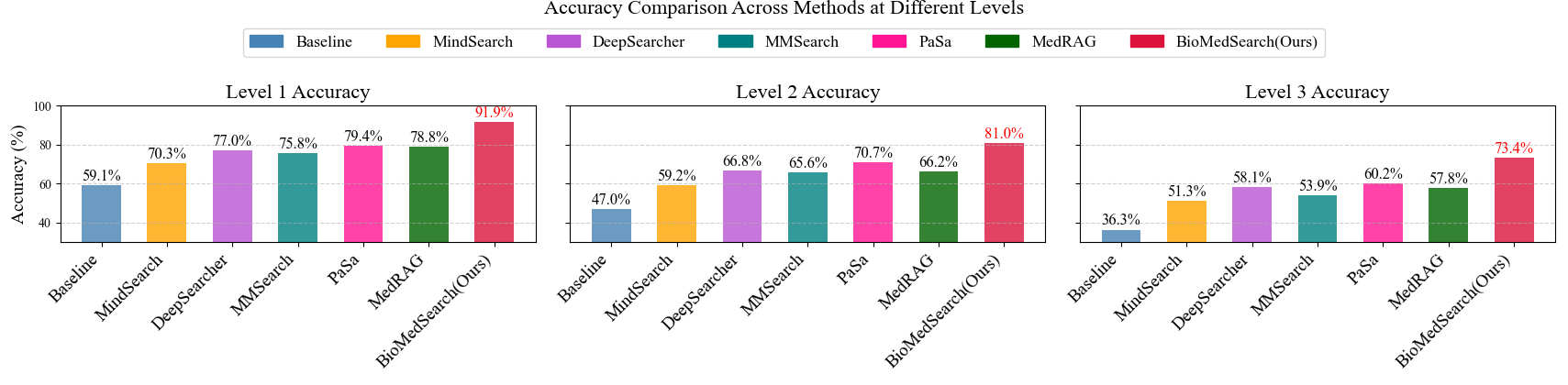}  
		\vspace{-5mm}
		\caption{Comparison of average accuracy across different methods on BioMedMCQs at three reasoning levels (Level 1–3). For each method, the reported value represents the mean accuracy across all LLMs at the corresponding level, reflecting overall performance in multi-level biomedical MCQs.}
		\label{fig:methods_compare}
	\end{figure*}
	\begin{table*}[!t]
		\setlength{\belowrulesep}{0pt}  
		\setlength{\aboverulesep}{0pt}  
		\centering
		\caption{Performance on BioMedMCQs Based on Random Biomedical Topic (\%)}
		\vspace{-2mm}  
		\setlength{\belowcaptionskip}{-6mm}  
		\footnotesize
		\renewcommand{\arraystretch}{0.9}  
		\footnotesize
		\resizebox{\textwidth}{!}{
			\begin{tabular}{p{3.4cm}p{2.0cm}>{\centering\arraybackslash}p{2.7cm}>{\centering\arraybackslash}p{2.7cm}>{\centering\arraybackslash}p{2.7cm}}
				\toprule
				\multirow{2}{*}{Methods} & \multirow{2}{*}{Models} & \multicolumn{3}{c}{Accuracy (\%) (1000 questions per level)} \\
				\cmidrule(lr){3-5}
				& & Level 1 & Level 2 & Level 3 \\
				\midrule
				\multirow{5}{*}{Baseline} 
				& ChatGPT-4.1~\cite{openai2025gpt4.1}         & 63.5 $\pm$ 0.7 & 53.8 $\pm$ 1.9 & 49.5 $\pm$ 0.5 \\
				& DeepSeek-R1~\cite{guo2025deepseek}    & 50.7 $\pm$ 1.2 & 39.7 $\pm$ 1.1 & 25.9 $\pm$ 1.0 \\
				& Llama-4~\cite{meta2025llama}    & 65.0 $\pm$ 1.5 & 57.4 $\pm$ 1.0 & 50.6 $\pm$ 0.9 \\
				& Gemini-2.5~\cite{comanici2025gemini}    & 61.6 $\pm$ 0.9 & 56.1 $\pm$ 0.8 & 42.7 $\pm$ 1.8 \\
				& Qwen3~\cite{yang2025qwen3technicalreport}       & 54.8 $\pm$ 2.1 & 41.8 $\pm$ 1.8 & 32.8 $\pm$ 0.7 \\
				\midrule
				\multirow{5}{*}{MindSearch~\cite{chen2024mindsearch}} 
				& ChatGPT-4.1~\cite{openai2025gpt4.1}         & 77.6 $\pm$ 1.1 & 69.8 $\pm$ 0.9 & 63.2 $\pm$ 2.0 \\
				& DeepSeek-R1~\cite{guo2025deepseek}    & 60.2 $\pm$ 1.0 & 54.7 $\pm$ 1.5 & 46.9 $\pm$ 0.9 \\
				& Llama-4~\cite{meta2025llama}    & 79.3 $\pm$ 2.4 & 68.8 $\pm$ 0.8 & 64.6 $\pm$ 0.7 \\
				& Gemini-2.5~\cite{comanici2025gemini}    & 73.9 $\pm$ 1.2 & 68.6 $\pm$ 0.6 & 57.3 $\pm$ 2.1 \\
				& Qwen3~\cite{yang2025qwen3technicalreport}       & 60.6 $\pm$ 1.1 & 53.2 $\pm$ 2.2 & 41.6 $\pm$ 0.7 \\
				\midrule
				\multirow{5}{*}{DeepSearcher~\cite{zheng2025deepresearcher}}
				& ChatGPT-4.1~\cite{openai2025gpt4.1}         & 81.4 $\pm$ 1.2 & 76.1 $\pm$ 1.1 & 71.2 $\pm$ 0.5 \\
				& DeepSeek-R1~\cite{guo2025deepseek}    & 75.2 $\pm$ 0.9 & 53.7 $\pm$ 0.6 & 42.0 $\pm$ 1.3 \\
				& Llama-4~\cite{meta2025llama}    & 83.9 $\pm$ 0.4 & 74.0 $\pm$ 1.1 & 68.8 $\pm$ 1.4 \\
				& Gemini-2.5~\cite{comanici2025gemini}    & 78.9 $\pm$ 1.1 & 72.3 $\pm$ 0.7 & 69.3 $\pm$ 0.5 \\
				& Qwen3~\cite{yang2025qwen3technicalreport}       & 65.5 $\pm$ 0.9 & 57.7 $\pm$ 0.4 & 49.3 $\pm$ 0.7 \\
				\midrule
				\multirow{5}{*}{MMSearch~\cite{jiang2024mmsearch}}
				& ChatGPT-4.1~\cite{openai2025gpt4.1}         & 79.6 $\pm$ 1.3 & 72.1 $\pm$ 2.2 & 63.2 $\pm$ 1.0 \\
				& DeepSeek-R1~\cite{guo2025deepseek}    & 71.7 $\pm$ 0.8 & 60.6 $\pm$ 1.5 & 45.0 $\pm$ 1.2 \\
				& Llama-4~\cite{meta2025llama}    & 82.3 $\pm$ 0.4 & 71.3 $\pm$ 0.7 & 64.9 $\pm$ 0.5 \\
				& Gemini-2.5~\cite{comanici2025gemini}    & 81.9 $\pm$ 0.6 & 69.7 $\pm$ 1.3 & 57.3 $\pm$ 0.9 \\
				& Qwen3~\cite{yang2025qwen3technicalreport}       & 63.6 $\pm$ 0.5 & 54.2 $\pm$ 0.7 & 49.2 $\pm$ 0.9 \\
				\midrule
				\multirow{5}{*}{PaSa~\cite{he2024pasa}}
				& ChatGPT-4.1~\cite{openai2025gpt4.1}         & 84.6 $\pm$ 1.7 & 76.5 $\pm$ 1.3 & 70.9 $\pm$ 1.6 \\
				& DeepSeek-R1~\cite{guo2025deepseek}          & 74.8 $\pm$ 0.6 & 68.3 $\pm$ 0.7 & 51.5 $\pm$ 2.2 \\
				& Llama-4~\cite{meta2025llama}      & 85.9 $\pm$ 1.4 & 78.5 $\pm$ 0.6 & 69.4 $\pm$ 1.0 \\
				& Gemini-2.5~\cite{comanici2025gemini}      & 83.8 $\pm$ 1.2 & 71.4 $\pm$ 0.7 & 63.4 $\pm$ 0.8 \\
				& Qwen3~\cite{yang2025qwen3technicalreport}          & 67.7 $\pm$ 0.9 & 59.0 $\pm$ 1.3 & 50.8 $\pm$ 1.0 \\
				\midrule
				\multirow{1}{*}{Self-BioRAG-7B~\cite{jeong2024improving}} 
				& - & 58.1 $\pm$ 0.9 & 47.8 $\pm$ 1.0 & 35.4 $\pm$ 2.3 \\
				\midrule
				\multirow{5}{*}{MedRAG~\cite{zhao2025medrag}} 
				& ChatGPT-4.1~\cite{openai2025gpt4.1}         & 87.2 $\pm$ 0.6 & 80.5 $\pm$ 0.9 & 72.8 $\pm$ 2.2 \\
				& DeepSeek-R1~\cite{guo2025deepseek}    & 74.9 $\pm$ 0.7 & 69.6 $\pm$ 0.8 & 62.5 $\pm$ 1.0 \\
				& Llama-4~\cite{meta2025llama}    & 86.2 $\pm$ 0.5 & 77.8 $\pm$ 1.4 & 63.9 $\pm$ 1.1 \\
				& Gemini-2.5~\cite{comanici2025gemini}    & 85.1 $\pm$ 2.3 & 78.7 $\pm$ 2.1 & 72.4 $\pm$ 1.3 \\
				& Qwen3~\cite{yang2025qwen3technicalreport}       & 72.9 $\pm$ 1.7 & 67.4 $\pm$ 1.5 & 60.3 $\pm$ 0.7 \\
				\midrule
				\multirow{5}{*}{BioMedSearch(Ours)} 
				& ChatGPT-4.1~\cite{openai2025gpt4.1}         & 93.8 $\pm$ 2.2 & 83.3 $\pm$ 1.2 & \textbf{78.0} $\pm$ \textbf{1.7} \\
				& DeepSeek-R1~\cite{guo2025deepseek}    & 89.9 $\pm$ 2.3 & 73.8 $\pm$ 1.4 & 67.7 $\pm$ 1.5 \\
				& Llama-4~\cite{meta2025llama}    & 92.1 $\pm$ 0.9 & \textbf{84.9} $\pm$ \textbf{1.3} & 74.1 $\pm$ 1.0 \\
				& Gemini-2.5~\cite{comanici2025gemini}    & \textbf{94.2} $\pm$ \textbf{1.0} & 82.6 $\pm$ 0.6 & 75.5 $\pm$ 1.3 \\
				& Qwen3~\cite{yang2025qwen3technicalreport}       & 89.4 $\pm$ 1.6 & 80.5 $\pm$ 1.4 & 71.8 $\pm$ 0.9 \\
				\bottomrule
				\vspace{-8mm}
			\end{tabular}
		}
		\label{tab:mcq_final_fitsingle}
	\end{table*}
	\setlength{\textfloatsep}{3pt}
	\begin{table}[htbp]
		\vspace{-3mm}  
		\setlength{\belowrulesep}{0pt}  
		\setlength{\aboverulesep}{0pt}  
		\centering
		\caption{BioMedMCQs Accuracy (\%) with Std. Dev.}
		\scriptsize
		\vspace{-2mm} 
		\scriptsize
		\setlength{\tabcolsep}{2pt} 
		\renewcommand{\arraystretch}{0.9} 
		\begin{tabular}{p{1.2cm} p{2.2cm} >{\centering\arraybackslash}p{1.5cm} >{\centering\arraybackslash}p{1.5cm} >{\centering\arraybackslash}p{1.5cm}}
			\toprule
			\multicolumn{1}{c}{Method} & \multicolumn{1}{c}{Model} & Level 1 & Level 2 & Level 3 \\
			\midrule
			\multirow{5}{*}{\centering \textbf{full setting}}
			& ChatGPT-4.1~\cite{openai2025gpt4.1}         & 93.8 $\pm$ 2.2 & 83.3 $\pm$ 1.2 & \textbf{78.0} $\pm$ \textbf{1.7} \\
			& DeepSeek-R1~\cite{guo2025deepseek}    & 89.9 $\pm$ 2.3 & 73.8 $\pm$ 1.4 & 67.7 $\pm$ 1.5 \\
			& Llama-4~\cite{meta2025llama}    & 92.1 $\pm$ 0.9 & \textbf{84.9} $\pm$ \textbf{1.3} & 74.1 $\pm$ 1.0 \\
			& Gemini-2.5~\cite{comanici2025gemini}    & \textbf{94.2} $\pm$ \textbf{1.0} & 82.6 $\pm$ 0.6 & 75.5 $\pm$ 1.3 \\
			& Qwen3~\cite{yang2025qwen3technicalreport}       & 89.4 $\pm$ 1.6 & 80.5 $\pm$ 1.4 & 71.8 $\pm$ 0.9 \\
			\midrule
			\multirow{5}{*}{\shortstack{w/o\\literature}}
			& ChatGPT-4.1~\cite{openai2025gpt4.1} & 87.4 $\pm$ 1.4 & 78.9 $\pm$ 1.3 & 72.1 $\pm$ 0.9 \\
			& DeepSeek-R1~\cite{guo2025deepseek}    & 70.3 $\pm$ 1.6 & 62.7 $\pm$ 0.6 & 56.6 $\pm$ 0.7 \\
			& Llama-4~\cite{meta2025llama}   & 85.9 $\pm$ 0.5 & 77.8 $\pm$ 0.6 & 70.2 $\pm$ 1.4 \\
			& Gemini-2.5~\cite{comanici2025gemini}   & 87.1 $\pm$ 1.3 & 76.3 $\pm$ 2.2 & 71.0 $\pm$ 1.0 \\
			& Qwen3~\cite{yang2025qwen3technicalreport}     & 81.0 $\pm$ 1.9 & 75.6 $\pm$ 1.2 & 67.9 $\pm$ 1.5 \\
			\midrule
			\multirow{5}{*}{\shortstack{w/o\\keywords}}
			& ChatGPT-4.1~\cite{openai2025gpt4.1} & 89.2 $\pm$ 0.9 & 80.8 $\pm$ 1.0 & 74.7 $\pm$ 1.3 \\
			& DeepSeek-R1~\cite{guo2025deepseek}    & 79.5 $\pm$ 2.1 & 71.6 $\pm$ 0.7 & 62.6 $\pm$ 0.6 \\
			& Llama-4~\cite{meta2025llama}   & 90.9 $\pm$ 1.2 & 79.2 $\pm$ 1.1 & 70.3 $\pm$ 1.0 \\
			& Gemini-2.5~\cite{comanici2025gemini}   & 88.3 $\pm$ 1.8 & 78.2 $\pm$ 1.9 & 73.3 $\pm$ 0.5 \\
			& Qwen3~\cite{yang2025qwen3technicalreport}    & 85.8 $\pm$ 1.6 & 72.1 $\pm$ 2.0 & 68.2 $\pm$ 1.1 \\
			\bottomrule
		\end{tabular}
		\label{tab:ablation_mcq_333}
	\end{table}
	\subsection{Implement Details}
	
	We implement BioMedSearch using Python 3.10.16 and integrate multiple mainstream LLMs as backends to comprehensively evaluate the performance of different model architectures in biomedical reasoning tasks. The experiment tested five LLMs: ChatGPT-4.1, DeepSeek-R1, Gemini-2.5, Llama-4, and Qwen3. All LLMs are accessed through their official APIs with default configurations, without any fine-tuning.
	
	Specifically, BioMedSearch extracts coherent paragraphs from structured markdown-formatted research reports to serve as contextual input for the models. Each question and its options are vectorized using the PubMedBert embedding model, and cosine similarity is computed between the question and each paragraph. The top‑$k$ most relevant paragraphs are selected to serve as input context, as shown in Figure~\ref{fig:Answering_MCQs}.
	
	During the answering phase, all LLMs use the same prompt template and are explicitly restricted from accessing any external knowledge; Answers must be generated solely based on the provided context. This process effectively standardizes model input across conditions, ensuring that evaluation results more accurately reflect each model’s reasoning ability and contextual understanding under constrained knowledge settings.
	\subsection{Main Results}
	
	Table~\ref{tab:mcq_final_fitsingle} presents the accuracy of different methods on BioMedMCQs. Overall, our method consistently outperforms the existing search agents and biomedical RAG across all model architectures and the three reasoning levels. This demonstrates our method's capability to establish semantic alignment between retrieval and question answering based on user input, as well as its robust adaptability across models and generalization in reasoning tasks.
	
	At Level 1, our method benefits from high-quality keywords extraction and precise literature matching, which result in better alignment of retrieved information. On this level, our method achieves the highest accuracy of 94.2\% on Gemini-2.5, followed by 93.8\% on ChatGPT-4.1 and 92.1\% on Llama-4. In contrast, MedRAG~\cite{zhao2025medrag} achieves a maximum of 87.2\% in ChatGPT-4.1, and PaSa~\cite{he2024pasa} achieves 85.9\% in Llama-4. These results indicate that our method achieves 6--8\% of improvement across models, reflecting stronger performance in basic factual reasoning and semantic alignment. At Level 2, our method benefits from high-confidence paragraph filtering based on vector semantic similarity, which enhances the semantic relevance between the selected passages and the sub-topics, thereby providing more inferentially supportive context for the LLM. In this setting, Llama-4 achieves the highest accuracy of 84.9\%, closely followed by Gemini-2.5 (82.6\%) and ChatGPT-4.1 (83.3\%). In comparison, MedRAG~\cite{zhao2025medrag} achieves up to 80.5\%, and MMSearch reaches 71.3\%. Our method achieves a 2--13\% advantage across models, highlighting improved contextual support in complex semantic integration. At the most challenging Level 3, thanks to our fine-grained modeling of information structure and semantic paths, the models maintain stable performance under high-complexity scenarios. Our method achieves 78.0\% accuracy on ChatGPT-4.1, with 75.5\% on Gemini-2.5 and 74.1\% on Llama-4. In comparison, MedRAG~\cite{zhao2025medrag} achieves a maximum of 72.8\%, while DeepSearcher~\cite{zheng2025deepresearcher} and PaSa~\cite{he2024pasa} score 71.2\% and 70.9\%, respectively. This demonstrates our superior ability to handle dynamic contexts and multi-level causal chains through more effective use of retrieved information and robust reasoning.

	To provide a clearer comparison of BioMedSearch and other methods across different levels, we present the accuracy results in Figure~\ref{fig:methods_compare}. As shown, in the level 1, BioMedSearch achieves an average accuracy of 91.8\%, significantly outperforming the representative biomedical RAG method MedRAG~\cite{zhao2025medrag} (81.2\%) and the typical general-purpose SearchAgent method PaSa~\cite{he2024pasa} (79.3\%). In the level 2, BioMedSearch reaches 81.02\%, showing a substantial improvement over MedRAG~\cite{zhao2025medrag} (74.80\%) and PaSa~\cite{he2024pasa} (70.74\%). For the most challenging level 3, BioMedSearch maintains strong performance with 73.42\%, clearly surpassing MedRAG~\cite{zhao2025medrag} (66.38\%) and PaSa~\cite{he2024pasa} (61.2\%). Overall, BioMedSearch consistently outperforms these representative methods across all three levels, demonstrating superior multi-level reasoning and information integration capabilities.
	\subsection{Ablation Study}
	To evaluate the contribution of different components in our method, we conducted ablation studies on three difficulty levels of BioMedMCQs. We designed two ablated variants: w/o literature refers to the removal of external literature retrieval, relying only on protein databases and web search to generate research reports for answering; w/o keywords remove keywords extraction, directly assigning retrieval tools to each sub-query without decomposition. Ours full models retain both literature retrieval and keywords extraction components.
	
	As shown in Table~\ref{tab:ablation_mcq_333}, removing either component leads to decreased accuracy across all models and difficulty levels. Notably, under the w/o literature setting, model performance drops significantly. For example, DeepSeek-R1 and Gemini-2.5 see their Level 3 accuracy fall from 67.7\% and 75.5\% to 56.6\% and 71.0\%, respectively, indicating that literature retrieval is critical for answering complex biomedical questions. Similarly, removing keyword guidance (w/o keywords) also results in noticeable performance degradation. For instance, Qwen3’s Level 3 accuracy drops from 71.8\% to 68.2\% without keywords guidance. This suggests that decomposing sub-queries into keywords enhances retrieval precision and helps avoid noisy or distracting information that may negatively influence the model's decision-making. Overall, our full model consistently outperform the ablated variants across all LLMs, validating the complementary value of literature retrieval and keywords extraction.
	\section{Conclusion}
	In this paper, we propose the BioMedSearch framework, which leverages LLMs and multi-source heterogeneous information retrieval to enhance biomedical question answering performance. By integrating literature retrieve, protein databases, and web search results, the method employs sub-query decomposition and keywords extraction strategies to construct a DAG, enabling efficient scheduling and integration of highly relevant information. This approach significantly improves the accuracy and interpretability of question answering results. Experimental results demonstrate that BioMedSearch outperforms existing methods on benchmark datasets, particularly exhibiting stronger robustness and generalization in complex tasks involving causal and temporal reasoning.
	
	Future directions include refining task graph structures and retrieval paths, as well as incorporating domain-specific knowledge graphs or multimodal data (e.g., biomedical images and pathways) to further improve performance on complex reasoning tasks.
	
	\bibliographystyle{IEEEtran}     
	\bibliography{references_main}        
\end{document}